\documentclass[11pt, a4paper, logo, onecolumn, copyright, nonumbering]{deepmind}

\usepackage{graphicx}

\usepackage[authoryear, sort&compress, round]{natbib}

\usepackage{amsmath}
\usepackage{amsfonts}
\usepackage{graphicx}
\usepackage{siunitx}
\usepackage[dvipsnames]{xcolor}
\usepackage{wrapfig}
\usepackage{caption}
\usepackage{subcaption}
\usepackage[export]{adjustbox}
\usepackage{makecell}

\usepackage{url}

\usepackage{algorithm}
\usepackage{algpseudocode}

\usepackage{hyperref}

\title{Controlling Commercial Cooling Systems Using Reinforcement Learning}

\correspondingauthor{jerryjluo@deepmind.com, paduraru@deepmind.com}

\author[*,1]{Jerry Luo}
\author[*,1]{Cosmin Paduraru}
\author[1]{Octavian Voicu}
\author[1]{Yuri Chervonyi}
\author[3]{Scott Munns}
\author[1]{Jerry Li}
\author[1]{Crystal Qian}
\author[1]{Praneet Dutta}
\author[1]{Jared Quincy Davis}
\author[2]{Ningjia Wu}
\author[2]{Xingwei Yang}
\author[2]{Chu-Ming Chang}
\author[2]{Ted Li}
\author[2]{Rob Rose}
\author[2]{Mingyan Fan}
\author[2]{Hootan Nakhost}
\author[2]{Tinglin Liu}
\author[3]{Brian Kirkman}
\author[3]{Frank Altamura}
\author[3]{Lee Cline}
\author[3]{Patrick Tonker}
\author[3]{Joel Gouker}
\author[3]{Dave Uden}
\author[3]{Warren “Buddy” Bryan}
\author[3]{Jason Law}
\author[1]{Deeni Fatiha}
\author[2]{Neil Satra}
\author[1]{Juliet Rothenberg}
\author[2]{Mandeep Waraich}
\author[1]{Molly Carlin}
\author[2]{Satish Tallapaka}
\author[1]{Sims Witherspoon}
\author[2]{David Parish}
\author[1]{Peter Dolan}
\author[2]{Chenyu Zhao}
\author[1]{Daniel J. Mankowitz}

\affil[*]{Equal contributions}
\affil[1]{DeepMind}
\affil[2]{Google}
\affil[3]{Trane}

\begin{abstract}

This paper is a technical overview of DeepMind and Google's recent work on reinforcement learning for controlling commercial cooling systems. Building on expertise that began with cooling Google's data centers more efficiently, we recently conducted live experiments on two real-world facilities in partnership with Trane Technologies, a building management system provider. These live experiments had a variety of challenges in areas such as evaluation, learning from offline data, and constraint satisfaction. Our paper describes these challenges in the hope that awareness of them will benefit future applied RL work. We also describe the way we adapted our RL system to deal with these challenges, resulting in energy savings of approximately 9\% and 13\% respectively at the two live experiment sites.

\end{abstract}

\begin{document}

\maketitle

\section{Introduction}

Heating, ventilation and air conditioning (HVAC) is responsible for a significant percentage of global CO$_2$ emissions. For example, space cooling alone accounts for around 10\% of the world's total electricity demand \citep{futureofcooling}. Increasing the efficiency of HVAC systems can therefore be an important tool for climate change mitigation. At the same time, the increasing availability of HVAC data collection and management systems make data-driven, autonomous, real-time decisions at scale an increasingly appealing avenue for improving efficiency. Building on prior work controlling the cooling systems of Google's data centers \citep{google-ai-data-center-2}, 
we used reinforcement learning (RL) to improve the energy efficiency of HVAC control in two commercial buildings.

There are several reasons why we believe reinforcement learning is a good fit for HVAC control problems. First, HVAC control is clearly a decision-making problem, with decisions including turning equipment on and off, and adjusting how hard to run each piece of equipment. There is also a natural reward function (energy use), in addition to multiple constraints that must be satisfied to keep the occupants comfortable and ensure safe system operations. The data required for training an RL agent can be obtained from widely deployed building management systems (BMS), and automated supervisory control can be implemented via cloud connected BMS that are becoming more prevalent. There is also an important sequential decision making aspect since actions can have long term effects. Finally, unlike Model Predictive Control (MPC), RL does not require a  detailed and comprehensive physics-based model to be developed, validated, and maintained for each building.

In this paper we detail our experience using RL to provide real-time supervisory setpoint recommendations to the chiller plant (an essential and energy-intensive part of many HVAC systems) in two commercial buildings. This work has presented us with an array of challenges, from common ones such as expensive and noisy data to more novel ones such as having different operating modes and multi timescale dynamics. In order to mitigate these challenges we employed a combination of general RL solutions and domain-specific heuristics. The resulting system was benchmarked against heuristics-based controllers provided by Trane, and showed a 9-13\% reduction in energy use while satisfying system constraints.

We believe that this paper contributes to the applied RL literature in several important ways. First, we show that we can use the same underlying RL algorithm for controlling the cooling systems of several different large commercial buildings, and we can improve upon industry-standard heuristics-based controllers in doing so. Second, we identify concrete examples of RL challenges and describe how these challenges were addressed in a real system. Finally, we discuss what might be needed to further scale RL for HVAC control, providing a potential blueprint for unlocking more efficiency and scaling to new use cases in the future.

\section{Background}

\subsection{Markov Decision Process}
A Markov Decision Process (MDP) \citep{sutton2018reinforcement} is defined by a $5$-tuple $\langle S,A,R,\gamma, P \rangle$ where $S$ is the state space, $A$ is the action space, $R\in [R_{min}, R_{max}]$ is a bounded reward function, $\gamma \in [0,1]$ is a discount factor and $P:S\times A \rightarrow S$ is a transition function that models the transition from a state $s \in S$ and action $a \in A$ to a new state $s' \in S$. A policy $\pi:S \rightarrow A$ is defined as a mapping from states to actions (or a probability distribution over actions in the case of a stochastic policy).

A solution to an MDP is an optimal policy $\pi^*$ that maximizes the expected return, which is defined as $\sum_{t=0}^T \gamma^t r_t$, where $T$ is the optimization horizon. The state value function $V^\pi(s)$ is defined as the long term value for executing policy $\pi$ from state $s \in S$. The action value function $Q^\pi(s,a)$ is defined as the value for executing action $a \in A$ in state $s \in S$ and thereafter following the policy $\pi$. A trajectory of length $N$ is defined as a sequences of states, actions and rewards $\tau = \langle s_0, a_0, r_0, s_1, a_1, r_1 \cdots r_{N-1}, s_N \rangle$.

\subsection{Policy iteration}
\label{sec:policy_iteration}

Policy iteration \citep{sutton98reinforcement} is a class of reinforcement learning algorithms that alternate between two steps called policy evaluation and policy improvement.

The policy evaluation step estimates the value function of the current policy $\pi$. The two main types of methods for policy evaluation are temporal difference (TD) learning and Monte Carlo evaluation. TD learning updates $Q^\pi(s_t, a_t)$ in the direction of
\[r_t + \gamma Q^\pi(s_{t+1}, \pi(s_{t+1}))   .\]
Monte Carlo policy evaluation updates $Q^\pi(s_t, a_t)$ towards 
\begin{equation} \label{eq:mc_target}
\sum_{k = t}^T \gamma^{k-t} r_k  .
\end{equation}
When $Q^\pi(s_t, a_t)$ is a neural network this amounts to solving a regression problem that minimizes the distance between $Q^\pi$ and the respective target.

The policy improvement step updates the policy so that for each state $s$ the new policy $\pi'$ selects the action $a$ that maximizes the action value function, or increases its probability of selecting this action in a stochastic policy. This can be written as
\[ \pi'(s) = \arg \max_a Q^\pi(s, a). \]

\subsection{Cooling systems for large commercial buildings}
\begin{figure*}[h]
    \includegraphics[width=\textwidth]{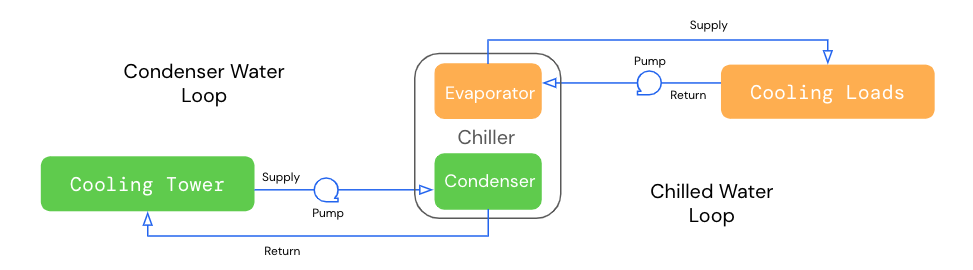}
    \caption{Schematic diagram of a water-cooled chiller plant with one chiller and one cooling tower. Note the distinct chilled and condenser water loops.
    }
    \label{fig:chiller_plant}
\end{figure*}

The fundamental problem that we solve is cooling the interior of large commercial buildings. The goal of the HVAC system is to maintain the air temperature and humidity in the building within a desired range for occupant comfort \citep{ashrae55_2020}, while providing adequate fresh air to maintain indoor air quality \citep{ashrae621_2020}. The cooling system reduces the interior air temperature by rejecting heat from the building into an external sink such as the atmosphere, and extracts water vapor from the interior air so it can be removed from the building. The total building load comes from incident solar radiation, convection from outdoor air, conduction from the ground, ventilation airflow, air infiltration and exfiltration, occupant metabolic processes, lighting, computers, and other equipment. It is measured in tons, which equals roughly 3.5kW. Due to the variety and variability of building loads, they are challenging to model, but it is possible using physics based models such as EnergyPlus™ \citep{energyplus}. However, the effort to create these models is substantial, and can be difficult to justify for older buildings without pre-existing models unless the building is going through major HVAC equipment upgrades. 

In larger facilities, it is common to utilize pumped water and fan-driven air as energy transport mechanisms. Typical components in the “water side” system include chillers, cooling towers, pumps, pipes, and control valves. Typical components in the “air side” system include air handlers, variable air volume units, fan coils, ductwork, dampers, and diffusers. Chilled water is used to distribute cooling to the section of the building where it is required. By circulating chilled water from the water side system and warm air from the HVAC spaces across water-to-air heat exchanger coils in the air side system, cooled and dehumidified air is provided to the HVAC spaces. The heat from the HVAC spaces is rejected into the water side system, increasing the temperature of the chilled water.

The chiller plant (Figure \ref{fig:chiller_plant}), part of the water side system, is responsible for cooling down the heated up water so that it can be reused to cool the air again. It cools the water by rejecting its heat into an external sink such as the atmosphere. With water-cooled chiller plants, this is typically done using cooling towers outside, which uses evaporative cooling. When the outside wet bulb temperature is not significantly colder than the temperature of the warm water, a piece of equipment called a chiller, typically utilizing a vapor-compression refrigeration cycle, is required to reject heat from the colder water to the warmer air. Chiller plants for larger buildings often contain multiple chillers, multiple cooling towers, and multiple water pumps, both to meet cooling capacity requirements and to provide redundancy. The chiller plant is controlled by the BMS. In our case this BMS was modified to allow for safely switching between obeying its normal heuristic-based controller, and accepting alternate recommendations provided by an external agent. Our RL control experiments were conducted on chiller plant systems.

\subsection{Sequence of Operations (SOO) for chiller control}
\label{sec:soo}
Most chiller plants are controlled using heuristics based on previous experience and first principle physics. These heuristics are typically programmed into a rule based control system called the \textit{Sequence of Operations} (SOO), which governs when to turn on/off various pieces of equipment and what setpoints to provide for them. Each \textit{setpoint} is the target value to be achieved by a closed-loop controller at the HVAC system level or at the equipment level. For example, each chiller will contain a setpoint for the chilled water temperature it will try to achieve. Because the combination of possible setpoints is massive, and because the optimal combination also depends on many external factors such as weather and building load, it can be very difficult and time consuming to design an SOO that provides close to optimal efficiency for all conditions. This is further complicated by the fact that the system may change over time due to changes in equipment conditions, weather patterns, load patterns, etc. In order to help alleviate these issues we propose using an RL policy, which is able to leverage the available data in order to learn and automatically determine a more efficient operating strategy.

\section{RL problem formulation}
\label{section:rl_problem_formulation}

We define our Chiller Control MDP as follows:

\begin{itemize}[noitemsep,nolistsep]
    \item Each state\footnote{As with most RL applications, we call this representation "state" even though the system is not necessarily Markovian with respect to it. We use the term "observation" and "state" interchangeably.} $s \in S$ is a numerical array of sensor measurement values such as temperatures, water flow rate, and equipment status (e.g. on/off). These are recorded at a 5 minute period, which naturally corresponds to a timestep in our MDP. We used feature engineering techniques and HVAC domain knowledge to select a subset of these measurements that are relevant, eventually settling on 50 state dimensions. More details about this can be found in Section \ref{section:submodels}.

\item Each action $a \in A$ is a 12 dimensional vector of real-valued setpoints available to the SOO, as well as discrete actions for turning on/off pieces of equipment. The full action space can be found in Appendix \ref{section:appendix_action_space}.

\item The reward $r_t$ is the negative of the total energy consumption of the chiller plant used within the 5 minute timestep $t$.

\item The agent solves a constrained optimization problem. The problem can be formalized as 
\begin{align}
\label{eq:main_objective}
\begin{split}
\arg \max_\pi \quad & E_\pi[\sum_{t=0}^T \gamma^t r_t] \\
\mathrm{s.t.} \quad & x \leq a_t \leq y \quad \forall t \in \{0 \dots T\}\\
& c_s(s_t, \dots s_{T_c}) \leq u
\end{split}
\end{align}
There are two sets of constraints. One is constraints on the actions, which can be guaranteed to be satisfied by the policy, and another is constraints on the observations, which require a prediction model. Here horizon $T$ is the value function (energy) horizon, $T_c$ is the observation constraint horizon, and $c_s(\cdot)$ is a function of a sequence of states (e.g. sum or max of the values of a particular sensor). We use different horizons for the value function and the observation constraints because the effects of actions on energy use can be manifested on a different time scale than their effects on some of the other observations.

\item The transition function is determined by the chiller plant behavior, the physical environment, the building load, and the local building management system (BMS). The BMS contains a set of safety checks to determine whether the action is safe, and can potentially modify the action before passing it to the equipment.

\end{itemize}

The RL control system uses a \textit{facility configuration} to store a representation of the state space, action space, objective, and constraints, as detailed in Appendix \ref{section:appendix_facility_configuration}.

\begin{minipage}{0.5\textwidth}
\begin{algorithm}[H]
\caption{Policy Improvement w/ Constraints}
\label{alg:policy_improvement}
\begin{algorithmic}
\For {each timestep $t$}
\State $A_t \gets$ sampled valid actions
\For {$a \in A_t$}
\State // Predict mean and std. dev. for energy 
\State // usage ($E$) and obs. constraint fn. ($c_s$)
\State $(\mu_E^a, \sigma_E^a, \mu_{c_s}^a, \sigma_{c_s}^a) \gets Q(s_t, a)$ 
\EndFor

\State // Filter out actions that are predicted to 
\State // violate obs. constraints
\State $A_t \gets \{a \in A_t | \mu_{c_s}^a + \alpha \sigma_{c_s}^a \leq u \}$

\State sample $x \sim \textrm{Uniform}(0,1)$
\If {$x \geq \epsilon$}
\State // Exploitation
\State $a_t \gets \arg \max_{a \in A_t} (-\mu_E^a - \alpha \sigma_E^a)$ \: 
\Else
\State // Exploration
\State $a_t \sim \textrm{softmax}_{a \in A_t}(-\mu_E^a + \alpha \sigma_E^a)$ \:
\EndIf
\State send $a_t$ as action to the facility
\EndFor
\end{algorithmic}
\end{algorithm}
\end{minipage}
\begin{minipage}{.05\textwidth}
\hspace{2cm}
\end{minipage}
\begin{minipage}{0.4\textwidth}
    \includegraphics[width=\textwidth]{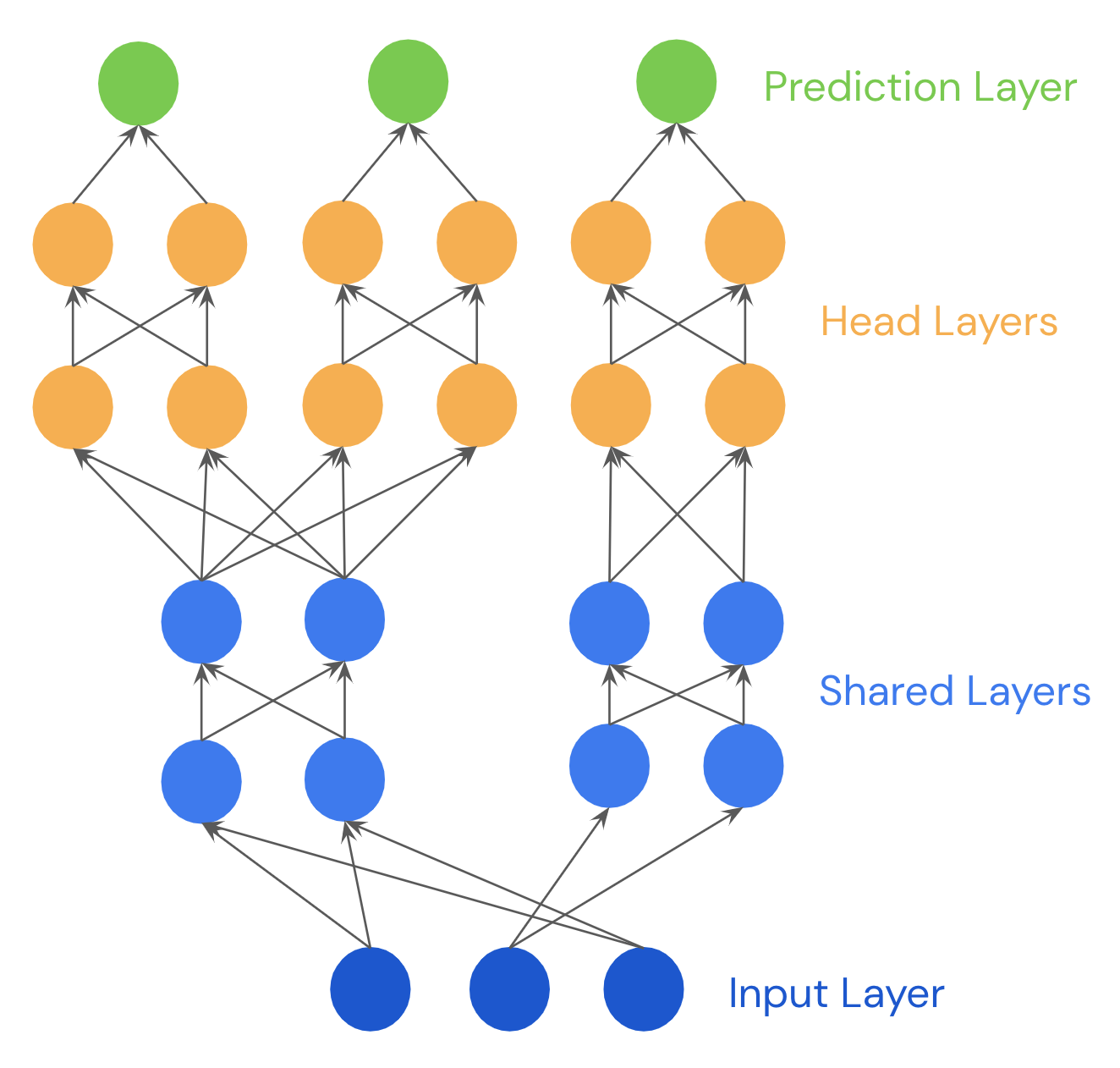}
    \captionof{figure}{Action value function architecture. The input layer are the (state, action) tuples. The shared layers are shared for all predictions within each tower. The head layers are specific for each prediction.}
    \label{fig:submodels_architecture}
\end{minipage}

\section{BVE-based COnstrained Optimization Learner with Ensemble Regularization}

The core RL algorithm that we focus on in this paper is similar to the one used for controlling the cooling in Google's data centers \citep{google-ai-data-center-2}. At a high level, the algorithm
\begin{itemize}[noitemsep,nolistsep]
    \item Learns an ensemble of neural networks for predicting the action value function and observation constraint violations. This is learned from offline behavior policy\footnote{The behavior policy is the policy that was used for generating the offline dataset.} data, similarly to BVE in \citet{gulcehre2021regularized}.
    \item Discards, at each time step, the actions that are known to violate the action constraints or predicted to violate the observation constraints.
    \item Selects one of the remaining actions based on an exploration-exploitation tradeoff involving the ensemble standard deviation.
\end{itemize}

However, unlike \citet{google-ai-data-center-2} our agent controls multiple chillers, which significantly increases the action space and the number of total constraints present. We also needed to deal with the fact that cooling demand can change quite frequently in a commercial building depending on occupancy level, whereas in a Google data center it tends to be relatively stable. Therefore, we made domain knowledge based additions to the core algorithm such as feature engineering, constraint enforcement heuristics, and action pruning, as discussed in Section \ref{section:challenges}. We call the resulting algorithm \textbf{BCOOLER}: BVE-based COnstrained Optimization Learner with Ensemble Regularization. We would like to highlight two important real-world aspects of our algorithm:
\begin{enumerate}[noitemsep,nolistsep]
    \item BCOOLER essentially performs a form of policy iteration directly on the real system: each day the current policy is re-evaluated using data up to and including the most recent day, and the next policy is produced by policy improvement using the updated $Q$ function. This means that we need an approach that is both conservative in order to work well when trained with offline data, but also somewhat exploratory in order to generate interesting data.
    \item We know that it is very important to respect the observation constraints in Equation \ref{eq:main_objective}, but some small amount of violation is permitted. However, we found it very hard to elicit specific penalty values for a soft constraint formulation, and impossible to guarantee that a hard constraint formulation as a function of future observations will be satisfied. Therefore we designed BCOOLER to satisfy the observation constraints if the prediction model is correct, with further manual tweaks used when necessary (See Sections \ref{section:adding-heuristics} and \ref{section:changing-constraints}).
\end{enumerate}

\subsection{Policy evaluation with constraints}
\label{policy_evaluation_with_constraints}

The policy evaluation step learns a multi-output action value function $Q \in \mathbb{R}^d$, where $d$ is the number of outputs, from real experience collected during previous building control. The action value function has one output corresponding to energy consumption and a separate output for each of the the observation constraints. We use Monte Carlo style targets for both the energy consumption and observation constraint predictions.

The targets for the energy consumption objective are generated by taking the time-normalized average power consumption in the trajectory from $t$ until $t+h$, where $h$ is a pre-configured look-ahead horizon. This is a modified version of the Monte Carlo target (\ref{eq:mc_target}) where $\gamma=1$ and the sum ends at $t+h$ instead of $\infty$. Each observation constraint target is generated by taking either the maximum or minimum value of the constrained observation in the trajectory from $t$ until $t+h$. Whether we take the maximum or the minimum depends on the sign in the observation constraint inequality (see Appendix \ref{section:appendix_facility_configuration}). A "$\leq$" sign implies taking the maximum, and vice versa.

 We decided to use Monte Carlo targets instead of TD style targets for several reasons:
 \begin{itemize}[noitemsep,nolistsep]
     \item We use a finite horizon problem formulation because it is easier to understand for facility managers than discounting, and \citet{amiranashvili2018td} find that "finite-horizon MC is not inferior to TD, even when rewards are sparse or delayed".
     \item The need to predict min and max values for the observation constraints makes TD style targets particularly challenging.
     \item One of the main advantages TD style methods have over Monte Carlo approaches is the ability to perform off-policy training. However, off-policy training is known to be problematic when RL is used with offline data \citep{levine2020offline, gulcehre2021regularized}. \item Our policy is based on a large action search (see Section \ref{sec:policy_improvement}), which makes TD-style off-policy updates computationally expensive.
 \end{itemize}

The $d$-dimensional action value function $Q$ is modeled by a multi-headed neural network (Figure \ref{fig:submodels_architecture}), whose inputs are (state, action) tuples, and whose labels are the objective and observation constraint targets described before. Due to the scarcity of data, feature selection and engineering is critical to creating good performing models. Since different heads in the model may have different optimal sets of input features, the neural network is divided into multiple towers, where each tower predicts a subset of the targets (see Section \ref{section:submodels} for further justification of this choice). The base layer to all the towers is the superset of all relevant input features, from which a mask selects a subset for each tower. This is followed first by shared fully connected layers, and then by separate fully connected layers for each prediction head. The labels for each target is normalized using their z-scores, and the final loss is the sum of the mean squared errors for all heads.

In order to give the agent a better understanding of its prediction uncertainty, we train an ensemble of neural networks \citep{lakshminarayanan2017uncertainty}. Every member of the ensemble has the same architecture, but is initialized with different randomized weights and receives the data in a different order. The standard deviation of the ensemble predictions is then used for controlling the exploration-exploitation trade off, as discussed in Algorithm \ref{alg:policy_improvement} and Section \ref{sec:policy_improvement}. 

In total the action value function has 62 input features (50 observations plus 12 actions) and 25 predictions (1 objective plus 24 observation constraints). A month of training data has around 10k examples. This model is retrained daily using all data collected up to that day.

\subsection{Policy improvement with constraints}
\label{sec:policy_improvement}

The policy improvement phase (see Algorithm \ref{alg:policy_improvement}), also referred to as control or inference, selects actions predicted to minimize expected energy usage over time while satisfying both action and observation constraints. This is done by first evaluating a large number (100k) of sampled valid actions using the action constraints (see Appendix \ref{section:appendix_facility_configuration}), where ``valid actions" means that the action constraints in Equation \ref{eq:main_objective} are respected. Next, actions for which the observation constraints in Equation \ref{eq:main_objective} are predicted to be violated\footnote{We add the ensemble standard deviation to the constraint prediction when evaluating whether the constraint is violated, in order to use a pessimistic estimate - see Algorithm \ref{alg:policy_improvement}} for the current state are filtered out. This ensures that, if our observation constraint prediction model is correct, then the observation constraints are respected. The remaining actions are scored using a combination of energy prediction and ensemble standard deviation, and this score is then used to select either an exploring or an exploiting action. This procedure is inspired by \citet{osband2016uncertainty}. More details can be found in Appendix \ref{sec:appendix_policy_improvement_details}.

\section{Live experiments}

\subsection{Setup}

We conducted live A/B tests on two different real facilities. The first facility was a university, where we controlled a chiller plant that provided cooling for several buildings on that campus. The second facility was a multi-purpose commercial building that contained a shopping center, restaurants, residential apartments, and a clinic. In order to properly conduct an A/B test between BCOOLER and Trane's heuristic controller in a single facility we need to account for the external factors that are outside of our control but strongly influence the results. For example, changes in the outside weather or building load will significantly change the amount of energy used by the building's HVAC systems, regardless of which agent was in control.

One method that we used was to alternate daily between BCOOLER and the SOO controlling the facility. This way we would get to see reasonably consistent weather between the two policies. Since the load approximately follows both diurnal and weekly cycles, having alternating days will allow both agents to see a roughly even distribution of load as well. In order to make sure that the effect of one agent doesn't carry over to the next agent under control, we removed from the analysis the 2 hours of data after each agent handover.

Another method to mitigate the effects of external factors is to normalize for them in the analysis. We discretized both the external temperature and the building load and only compare hourly performance between BCOOLER and the SOO that fall into the same temperature and load bucket. Finally we will take an average of the performance difference between BCOOLER and the SOO weighted based on the amount of time spent in each temperature and load bucket.

Each A/B test was run over the course of 3 months, which took place in the shoulder season transitioning between cooler and warmer temperatures. This means that the agent was subject to a variety of outside weather conditions, which gave us more confidence that BCOOLER will generalize across the year.

\begin{figure*}[h]
    \begin{center}
    \includegraphics[width=\textwidth]{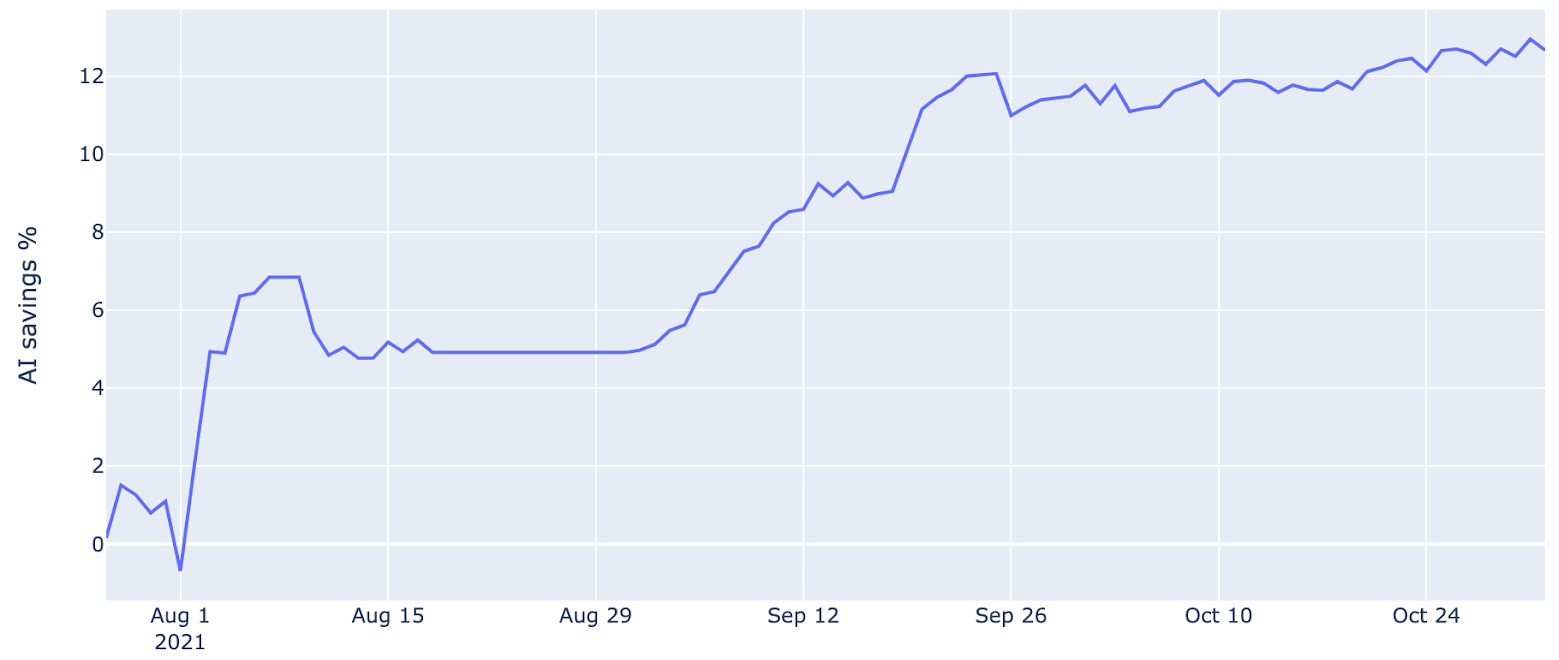}
    \end{center}
    \caption{Energy savings results over time for one of the live experiments. Each point represents the cumulative normalized savings since the beginning of the experiment. Note that the flat region in August was due to a pause in the experiment.}
    \label{fig:cumulative_savings}
\end{figure*}

\subsection{Results}

\begin{figure*}[h]
    \includegraphics[width=\textwidth]{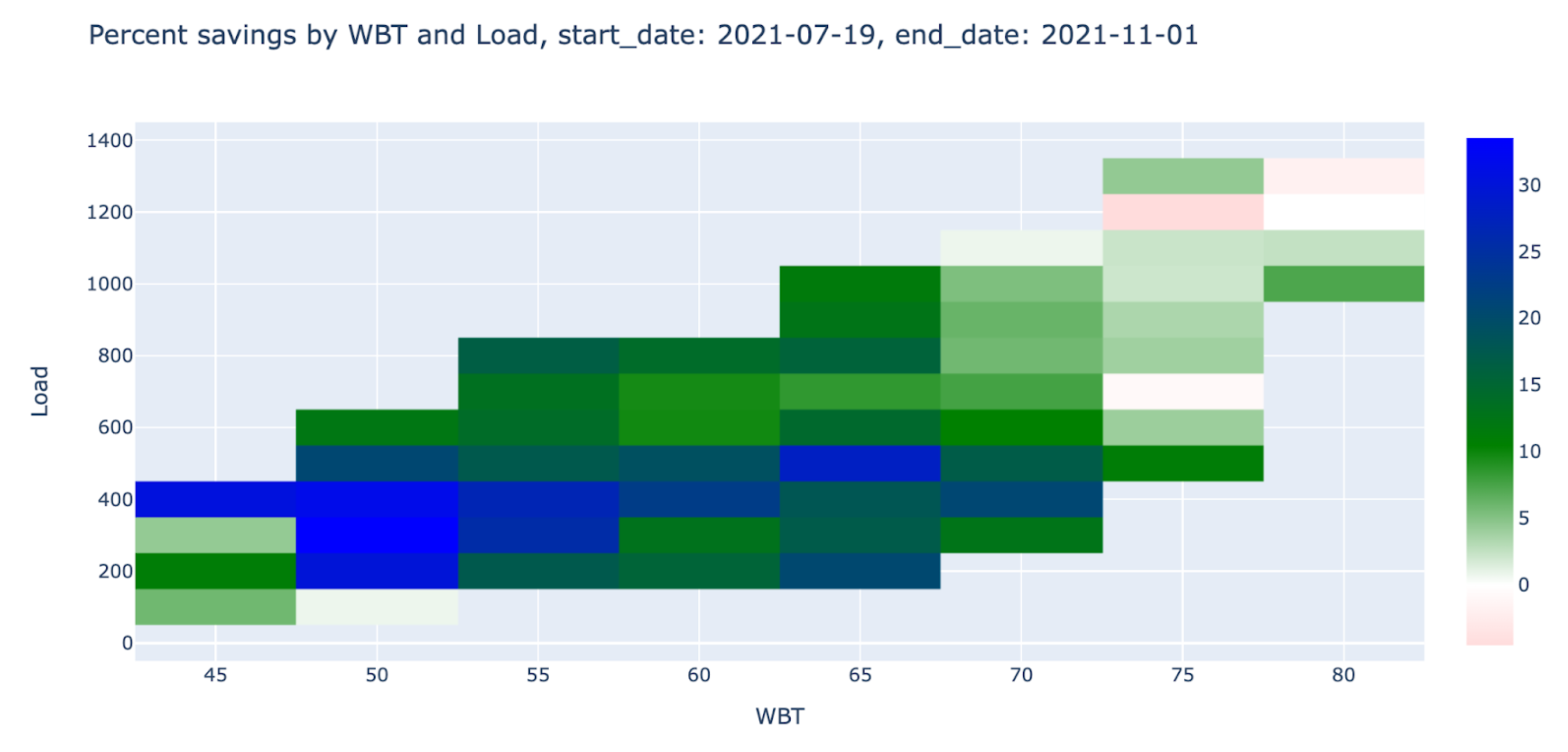}
    \caption{Energy savings results across different external conditions. Each box shows the percent energy savings (higher is better) BCOOLER was able to achieve for a combination of the outside wet-bulb temperature ($^\circ$F) and the building load (tons). We see that in general BCOOLER performs better at cooler temperatures and at lower loads.}
    \label{fig:savings_results}
\end{figure*}

\begin{figure*}[h]
    \includegraphics[width=\textwidth]{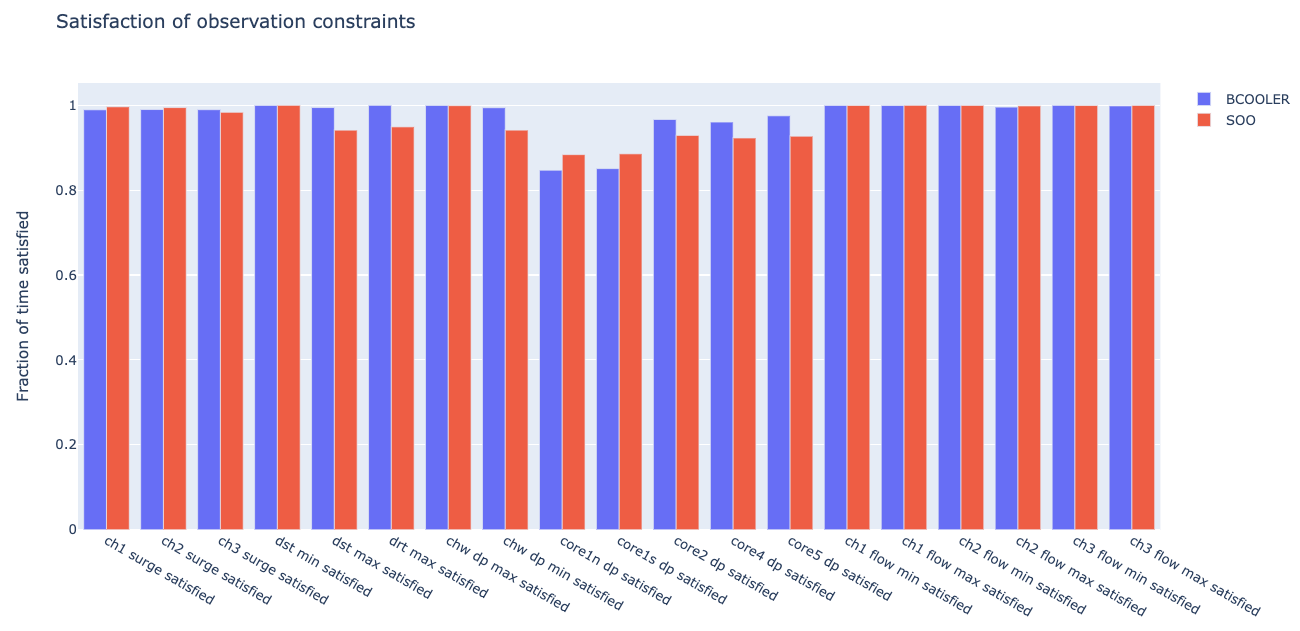}
    \caption{Results for how well the agent satisfied each observation constraints averaged over the course of one of the live experiments. Here we see that BCOOLER satisfied the observation constraints most of the time at a comparable rate to the SOO. There were times where the observation constraints were made too tight and thus both the AI and the SOO were violating the defined constraints. We worked together with Trane to fix them (See Section \ref{section:changing-constraints}). The fact that the SOO also sometimes violates the defined constraints further show how difficult it is to come up with an appropriate set of constraints for an AI.}
    \label{fig:constraint_satisfaction}
\end{figure*}

Across the duration of the live experiment BCOOLER was able to achieve 9\% and 13\% energy savings on the two live A/B tests against Trane's heuristics-based controllers (See Figure \ref{fig:cumulative_savings}), while still maintaining the same comfort levels of the occupants by respecting the constraints set by the building managers (See Figure \ref{fig:constraint_satisfaction}). One thing to note is that as part of the retro-commissioning effort to make the facilities "AI ready", the SOO was tuned to improve its performance. This implies that BCOOLER's performance improvement against an average chiller plant in the wild is likely higher than what we saw during these experiments.

Figure \ref{fig:cumulative_savings} shows that BCOOLER's performance improves over time. Hypotheses for why this is the case include BCOOLER learning a better model with more data, software fixes during the evaluation period, and the weather conditions allowing more room for optimization in the fall.

By breaking the energy savings down by the external factors such as temperature and building load (see Figure \ref{fig:savings_results}), we can see that BCOOLER performs better in some conditions than others. For example, BCOOLER does very well when the outside temperature is colder and when the load is lower, and conversely it does doesn't do quite as well when the outside temperatures is higher and when the load is higher (though it still does better than the SOO). One explanation for this phenomenon is that when the temperature and load is high, the equipment are running close to their max capacity, and there is less room for BCOOLER to make intelligent decisions, for example about the trade off between the utilization of certain pieces of equipment (see Appendix \ref{section:appendix_ai_learnings}).

We also identified specific behaviors that allowed BCOOLER to improve upon the baseline. One example is producing colder condenser water than the SOO in some situations, which requires more power use in the cooling towers but reduces overall power use of the plant by allowing the chillers to run more efficiently. BCOOLER also learned a policy that was robust to some amount of sensor miscalibration, learning to internally recalibrate them. More details on these behaviors can be found in Appendix \ref{section:appendix_ai_learnings}.

\begin{center}
\begin{table*}
\centering
\begin{tabular}{|| c | c ||} 
 \hline
 \textbf{Challenge} & \textbf{Mitigations} \\
 \hline\hline
 Learning and evaluation using limited data (\ref{section:challenge-limited-samples}) & \makecell{Sensitivity analysis and model unit tests (\ref{section:sensitivity-analysis}) \\ Feature engineering and submodels (\ref{section:submodels}) \\  Training on AI only data (\ref{section:training_ai_only}) \\ Data examination and cleaning (\ref{section:data_cleaning})} \\
 \hline
 \makecell{Abiding by complex constraints (\ref{section:challenges-constraints})} & \makecell{Hierarchical chiller heuristic (\ref{section:adding-heuristics}) \\ Tweaking the constraint specification (\ref{section:changing-constraints}) } \\
 \hline
  \makecell{Non-stationary dynamics and observations (\ref{section:noisy_tasks})} & Data examination and cleaning (\ref{section:data_cleaning}) \\
 \hline
 \makecell{Inference that must happen in real time (\ref{section:challenge_realtime})} & Action pruning (\ref{section:action_pruning}) \\
 \hline
 Different operation modes (\ref{section:different_operation_modes}) & Mode-specific action masking (\ref{section:action_masking}) \\
 \hline
 Multiple time scales for action effects (\ref{section:challenge-multi-timescale}) & Hierarchical chiller heuristic (\ref{section:adding-heuristics}) \\
 \hline 
\end{tabular}
\vspace{.2cm}
\caption{Some of the challenges we encountered when deploying BCOOLER, matched with the approaches we used for mitigating them. For each challenge and mitigation we point to the section that expands on it.}
\label{tab:challenges_mitigations}
\end{table*}
\end{center}

\section{Challenges and mitigations}
\label{section:challenges}

Applying RL to physical systems presents numerous challenges that are typically not encountered in simulated environments. Table \ref{tab:challenges_mitigations} describes some of the challenges we faced during BCOOLER's operation and the mitigations that we put in place to deal with them. We briefly outline them in this section, but much more detail on them can be found in Sections \ref{section:challenges_details} and \ref{section:mitigations_details}.

The challenges we faced can be summarized as:
\begin{enumerate}[noitemsep,nolistsep]
    \item \textbf{Learning and evaluation using limited data.} We had to learn directly from system interactions without access to a simulator. This made it very difficult to reliably evaluate model improvements, and also limits the diversity of behaviors that could be observed by the agent during training. (See Section \ref{section:challenge-limited-samples})
    \item \textbf{Abiding by complex constraints.} Due to the difficulty of always predicting which actions will violate the observation constraints we had to add heuristics to our agent to further reduce the amount of constraint violation. (See Section \ref{section:challenges-constraints})
    \item \textbf{Non-stationary dynamics and observations.} Weather patterns, load patterns, equipment performance, sensor calibration, can all change over time. In conjunction with limited data, this makes learning and inference more difficult. In particular, it means that for evaluating any model ablation we would need to run two different agent policies side by side for a long enough time, which was infeasible during the live experiments. (See Section \ref{section:noisy_tasks})
    \item \textbf{Inference that must happen in real time.} Because the presence of complex constraints limits the types of RL approaches we can use, generating actions fast enough for real time control can be difficult. (See Section \ref{section:challenge_realtime})
    \item \textbf{Different operation modes.} The system's action space, constraints, and dynamics can be quite different depending on the weather conditions, which makes control more challenging. (See Section \ref{section:different_operation_modes})
    \item \textbf{Multiple time scales for action effects.} Some actions immediately affect the system and have short term effects, while others have long lasting effects. Being able to account for both types of actions with a single agent is difficult. (See Section \ref{section:challenge-multi-timescale})
\end{enumerate}

Note that more generic forms for several of our challenges have been previously identified in other applied RL literature \citep{challenges}.

Some of the approaches we used to mitigate these challenges are listed below. Due to the limited duration of the live expriments and the variability in external factors it was impossible to assess how impactful each mitigation has been, however we do explain our reasoning for implementing them in detail in Section \ref{section:mitigations_details}.
\begin{enumerate}[noitemsep,nolistsep]
    \item \textbf{Sensitivity analysis and model unit tests.} We used domain understanding to develop unit tests that the action value function should abide by, in order to be more confident about evaluating the agent's performance prior to deployment. (See Section \ref{section:sensitivity-analysis})
    \item \textbf{Feature engineering and submodels.} We performed per-head feature engineering for the action value function in order to make sure that it works well for both the energy prediction and the observation constraint predictions. (See Section \ref{section:submodels})
    \item \textbf{Training on AI only data.} We favored the richness of the data generated by BCOOLER over the more static data generated by the SOO. (See Section \ref{section:training_ai_only})
    \item \textbf{Hierarchical heuristic policy.} We created a hierarchical policy with heuristics in order to handle actions with different effect time horizons. Actions with longer time horizons were governed by a high level policy, while actions with shorter time horizons were delegated to a lower level policy. (See Section \ref{section:adding-heuristics})
    \item \textbf{Tweaking the constraint specification.} We spent significant time working together with Trane to come up with a set of constraints that allows the agent to satisfy the customer's needs, while at the same time giving the agent ample room for optimization. (See Section \ref{section:changing-constraints})
    \item \textbf{Action pruning.} To combat the curse of dimensionality in the action space, we created an efficient way to prune and search through the action space. (See Section \ref{section:action_pruning})
    \item \textbf{Mode-specific action masking.} We masked various actions depending on the current state of the environment, allowing a single agent to handle multiple weather dependent modes that have different action spaces and constraints. (See Section \ref{section:action_masking})
    \item \textbf{Data examination and cleaning.} We spent significant time ensuring that the data was of high quality, both for historical data that was used to train the agent, and also live data that was used during control. (See Section \ref{section:data_cleaning})
\end{enumerate}

\section{Challenges details}
\label{section:challenges_details}
\subsection{Learning and evaluation using limited data}
\label{section:challenge-limited-samples}

During the time of the project we didn't have access to a high fidelity simulation that accurately captures the dynamics of the chiller plant that we are controlling. Because of this we can only resort to using data collected from the facility itself. There were 2 main sources of data: historical data collected by the SOO and data collected by BCOOLER as it is controlling the facility.

The historical data consists of less than a year of facility data from the SOO controlling the system. This is theoretically a good dataset to learn from, and can be thought of as a behavioral dataset from an expert policy. However, since the SOO is a hard-coded policy that is fully deterministic, many of the actions in the dataset were the same throughout the entire dataset. This makes learning difficult because the models won't be able to extrapolate well to the other actions that the SOO has not considered. This also makes evaluation difficult because having good prediction performance on the SOO's actions doesn't necessarily translate to good prediction performance on BCOOLER's actions. 

One phenomenon that we found was that when we conduct sensitivity analysis on the model's predictions with respect to the input actions, the model would often ignore the actions completely, relying purely on the state to make the prediction. Since most of the data had static actions, this results in a good loss for the model anyways. However, during control the agent heavily relies on the predicted effects of various actions in order to make a good decision. Both how we verified and solved this issue is discussed in Sections \ref{section:sensitivity-analysis} and \ref{section:submodels}.

The AI control data, on the other hand, contains rich exploration data that span a wide range of actions and states. This is the ideal data to train the agent with. However, due to there only being a single real environment, data collection is slow. The system collects data at 5 minute intervals, so each day will only produce around 300 samples. This is very limiting considering that the facility manager expects the system to be at parity with the SOO within a few weeks. 

Furthermore, even if the agent works well in the present, non-stationarities in the environment require the agent to be able to quickly make adaptations when conditions change (see Section \ref{section:noisy_tasks}).

\subsection{Abiding by complex constraints}
\label{section:challenges-constraints}

A major difficulty is translating the system constraints known by the facility manager into a language that BCOOLER can interpret and make use of. It is easy for the facility manager to look at a particular situation and say whether that situation violates any constraints, but it is very difficult to fully enumerate all the possible constraints that the system may have a priori, especially in a mathematically rigorous way that an AI agent can utilize. Through our previous experience as well as numerous discussions with Trane and the facility managers, the agent that we built had to respect 59 different constraints on the action space and 24 different constraints on the observation space.

The constraints on the action space had to be guaranteed to never be violated, and were of the form described in Appendix \ref{section:appendix_facility_configuration}. One challenge was that the action constraints sometimes depended on custom values that the facility manager picked out, and since these constraints can interact with each other in complex ways, there were instances where the combination of action constraints resulted in no feasible actions. Debugging this requires working together with the facility manager to understand which constraints are actually necessary, and whether some aspects of the action space were over-constrained (see Section \ref{section:changing-constraints}). This is an error prone process, especially due to the fact that the facility managers never previously needed to express constraints in such rigorous ways.

The constraints on the observation space didn't need guarantees against violation. However, they are more complex to satisfy because they require foresight from the agent that its current actions won't lead to violations in the future. What made this particularly challenging was the fact that some observation constraints required only 1 step planning to avoid, while others required multiple steps. See Section \ref{section:challenge-multi-timescale} for more details.

Separately, there are some observation constraints whose form is very complex. For example, one of the constraints that we saw had the form $t_1 \leq \min(h_3, 3+\max(h_1, \min(l_1, l_1 + \frac{h_1-l_1}{h_2-l_2}(o_1-l_2))))$, where $t_1$ is a temperature measurement, and $h_1,h_2,h_3,l_1,l_2,o_1$ are a mixture of measurements from the environment and facility manager defined values that may change over time.

Another issue with this complex combination of constraints is that it is difficult to build a policy network whose outputs satisfy them all. Doing so would require careful tuning of the reward trade-offs of the different objectives and constraints. This is one of the reasons why we chose the action space search approach outlined in Section \ref{sec:policy_improvement} instead of using an actor-critic approach that leverages a policy network.

\begin{figure*}[h]
    \includegraphics[width=\textwidth]{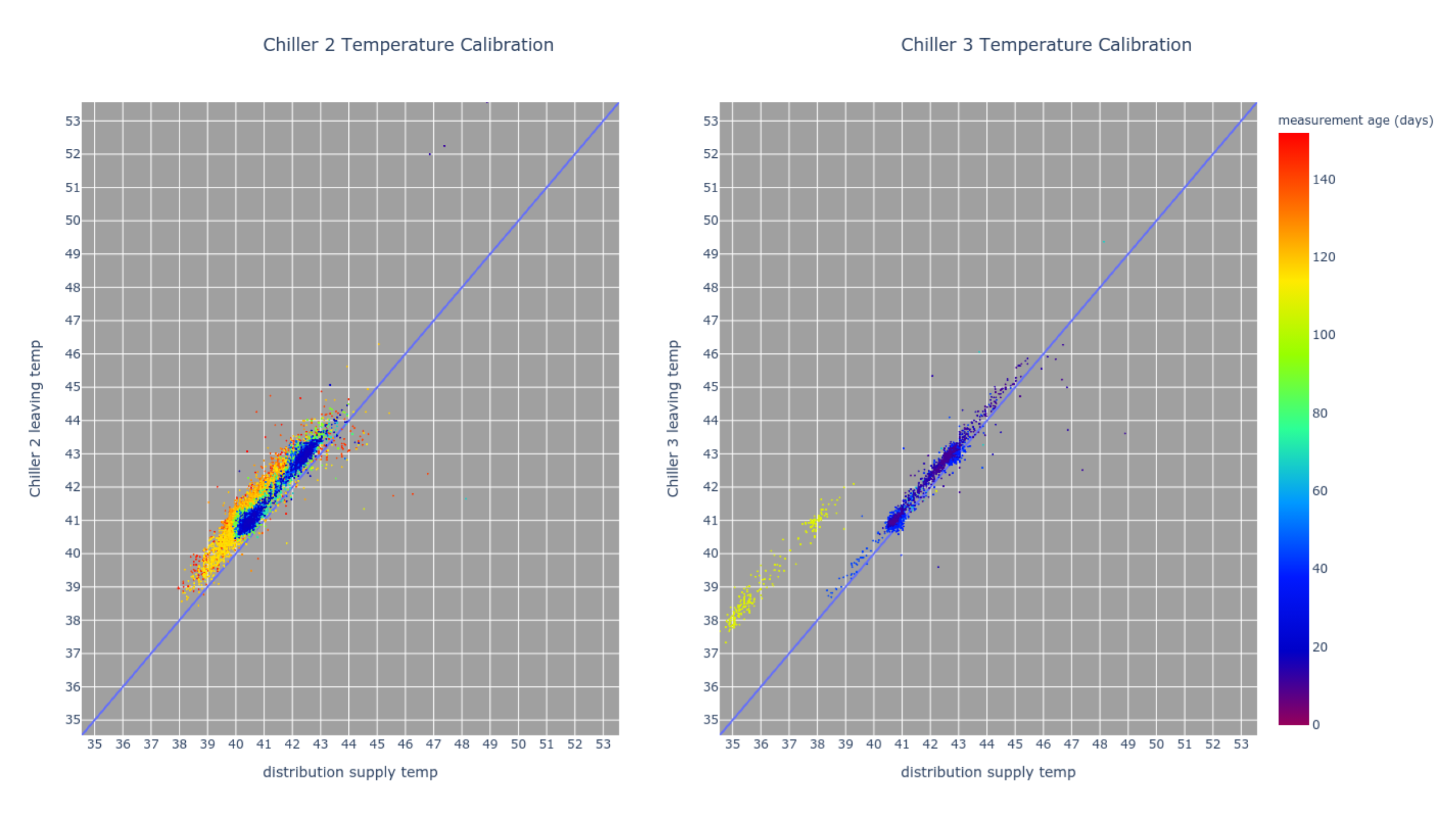}
    \caption{An example of sensor drift. The output of all chillers feed into a common water pipe, and the water temperature in that common pipe is called the distribution supply temperature. When a single chiller is running, the output temperature of the chiller is equivalent to the distribution supply temperature. In the graphs above we plot the output temperature of the chiller vs the distribution supply temperature when a single chiller is running. We expect all the points to be on the $y=x$ axis if the temperature sensors from the chillers and the common pipe are perfectly calibrated. However, we see above that this is not always the case. On the graph to the right, we see that at some point the chiller temperature sensors were calibrated, which resulted in a large calibration jump. One thing to note is that even with perfect calibration these sensors may still read differently because they are measuring water temperatures at slightly different places in the facility.}
    \label{fig:sensor_drift}
\end{figure*}

\subsection{Non-stationary dynamics and observations}
\label{section:noisy_tasks}

The environment that the agent is deployed to is non-stationary, which makes learning with limited data challenging. This means that the agent must be able to adapt to these changes very quickly. This is also another reason why the presence of historical data isn't sufficient to train a good agent.

One example of this non-stationarity is that the performance of the equipment in the facility changes over time. In chiller plants, the chiller’s capacity and performance degrades over a timescale of months to years, due to fouling of the heat exchanger surfaces by mineral scaling, biofilms, dust and dirt, and corrosion. Once the heat exchanger is cleaned, the chiller's efficiency immediately increases.

Another example is the presence of sensor drift. Over time, the temperature sensors in the facility will drift from their initial calibrated settings. Similar to the heat exchanger cleaning situation, if these sensors are recalibrated then there is again a large change in the distribution of sensor values (see Figure \ref{fig:sensor_drift}). Similarly, the implementation of the controller that reads the agent's actions and sends the raw commands to the equipment can change over time if improvements are identified and implemented. During the course of the live experiment, this was done several times.

These issues make it difficult to perform reliable evaluation, as any policy needs to be run for long enough to be tested under a variety of external conditions. This is the main reason that we do not report any model ablations in this paper - they were infeasible during the limited duration of the live experiments.

\subsection{Inference that must happen in real time}
\label{section:challenge_realtime}
Since control happens at a 5 minute period, the agent needs to be able to make a decision using fresh observations within 1 minute. Because of the complex constraints (see Section \ref{section:challenges-constraints}), the BCOOLER algorithm required generating and scoring a large number of candidate actions (see Section \ref{sec:policy_improvement}). This set of candidate actions grows exponentially with respect to the number of action dimensions, and is highly sensitive to the number of candidate values per action dimension. Without careful consideration for the size of the action space and how to carefully prune them, the control iteration can take a prohibitively long time.

\begin{figure*}[h]
    \includegraphics[width=\textwidth]{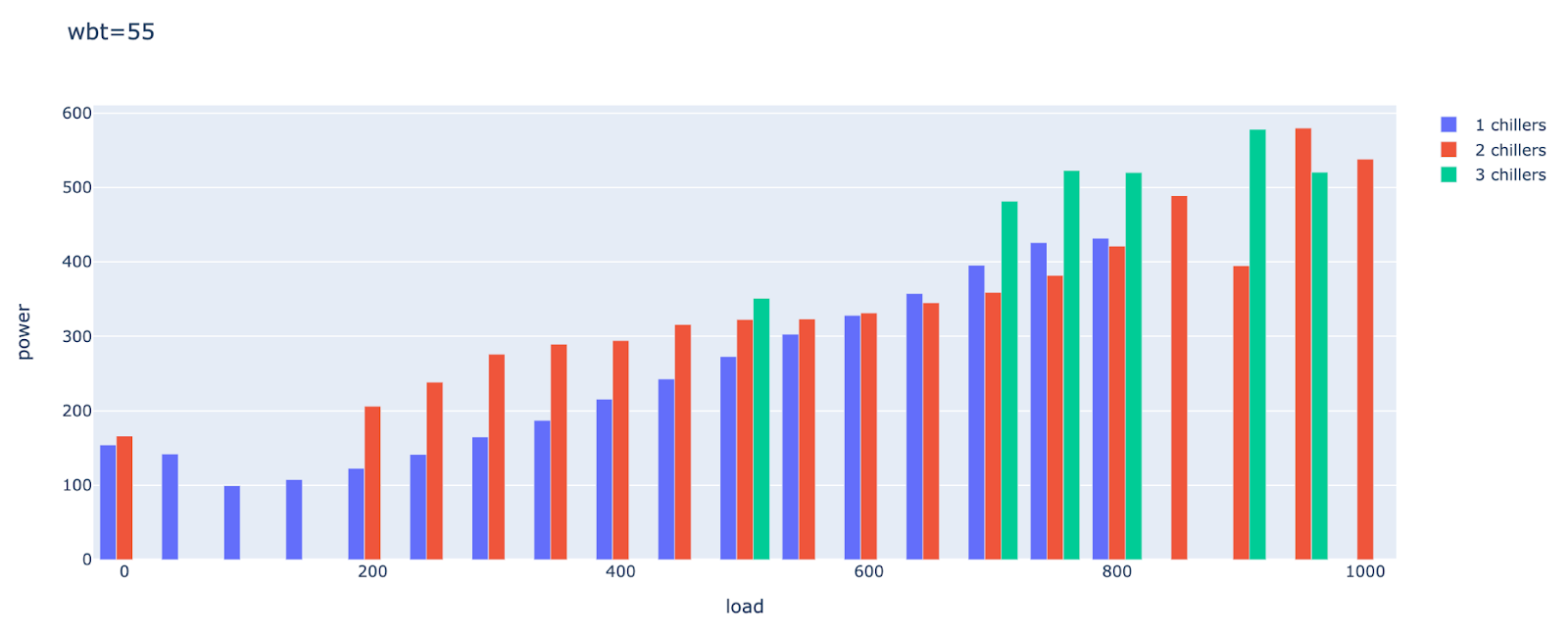}
    \caption{Total chiller power usage (kW) as a function of the building load (tons), looking only at the data where the outside wet bulb temperature was between 55$^{\circ}$F and 60$^{\circ}$F. We see that when the load is around 200 tons, using 2 chillers requires more energy than using 1 chiller. However, when the load is 800 tons, using 2 chillers requires less energy than using 1 chiller.}
    \label{fig:power_vs_n_chillers}
\end{figure*}

\subsection{Different operation modes}
\label{section:different_operation_modes}

Some chiller plants are capable of running in two distinct operating modes. The mechanical cooling mode, used in all vapor-compression chiller plants, is where the chiller plant runs the chillers and their refrigeration cycle using motor-driven compressor(s) in order to cool the facility when the outside temperatures are not very cold. Some chiller plants are also capable of running in another mode, called the free cooling mode or the water-side economizer mode. In this mode, the chiller plant cools the water passively without running the chiller's compressor(s) when the outside temperatures are cold enough. These two modes have different action spaces and constraints, and thus it is difficult for a single agent to be able to control both modes well. Another action that the agent has access to is figuring out when to switch between mechanical cooling and free cooling. To keep things simple, our agent handles the superset of all actions available during both modes, and ignores the actions that aren't relevant to a particular mode.

\begin{figure*}[h]
    \includegraphics[width=\textwidth]{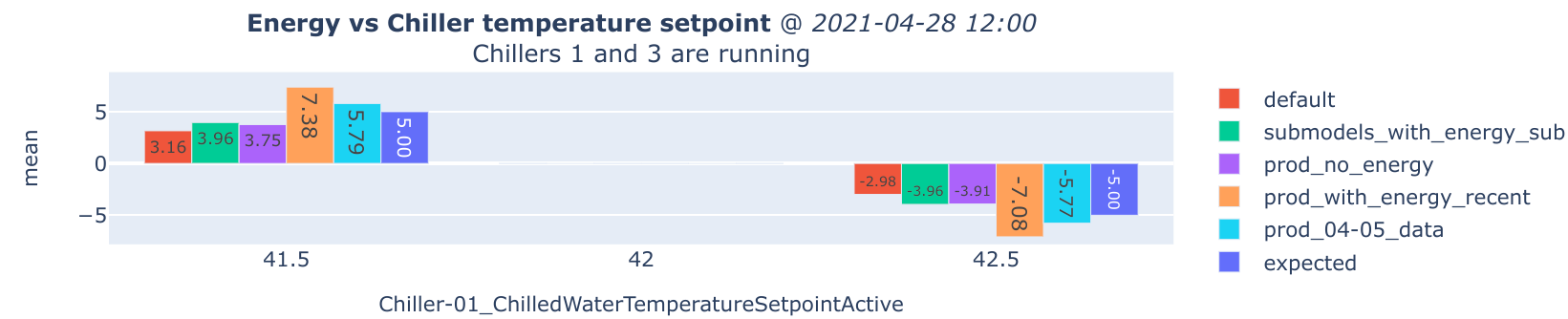}
    \caption{Example of an individual unit test, which is used to compare the model's sensitivity to changes in the  action features against what we expect them to be (labelled "expected"). Here we look at the relationship between energy usage and the chiller temperature setpoint. We pick a particular state where the chillers have a 42$^{\circ}$F temperature setpoint. We then ask the model to predict the relative change in the energy usage when we change the temperature setpoint by 0.5$^{\circ}$F up or down. In this particular case, we see that the model labelled as "prod\_04-05\_data" agrees with the "expected" value the most.}
    \label{fig:individual_unit_tests}
\end{figure*}

\subsection{Multiple time scales for action effects}
\label{section:challenge-multi-timescale}

The actions that we considered controlling for the chiller plant ranged in how long their effects influence the future state. For example, the action to change the flow rate of the water can be met very quickly, within the order of 5 minutes. If the flow rate setpoint is changed, the effects from the previous value of the setpoint is removed very quickly. Thus for observation constraints regarding the flow rate, the value function doesn't need to predict very far into the future in order to be accurate. On the other hand, another possible action was to control the sequence in which the chillers turned on and off when controlling the number of chillers to run. This was something that can only be changed every few weeks, and there are constraints on balancing how much each chiller runs proportionally over the course of several weeks. Controlling this action required being able to predict weeks into the future in order to make sure that constraints are satisfied. We ended up not controlling this action during the live experiment since its timescale was far too different from the others.

Because different actions have different time scales for their effects, a model trained on just predicting a specific future horizon for all targets may not do very well. Also, different horizons may require different sets of input features to incorporate the relevant dynamics for predicting that horizon.

\section{Mitigations details}
\label{section:mitigations_details}
In this section we describe how some of the previously mentioned issues were mitigated. One important theme is that these mitigations were only realized because we spent considerable time understanding the domain, both through discussions with HVAC experts 
and through the experience of deploying the agent.

\subsection{Sensitivity analysis and model unit tests}
\label{section:sensitivity-analysis}

\begin{figure*}[h]
\includegraphics[width=\textwidth]{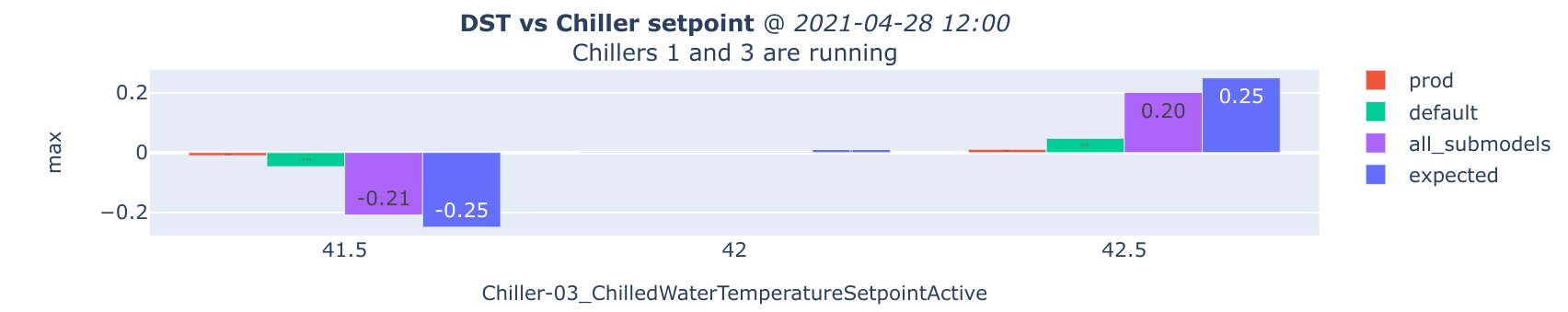}
\includegraphics[width=\textwidth]{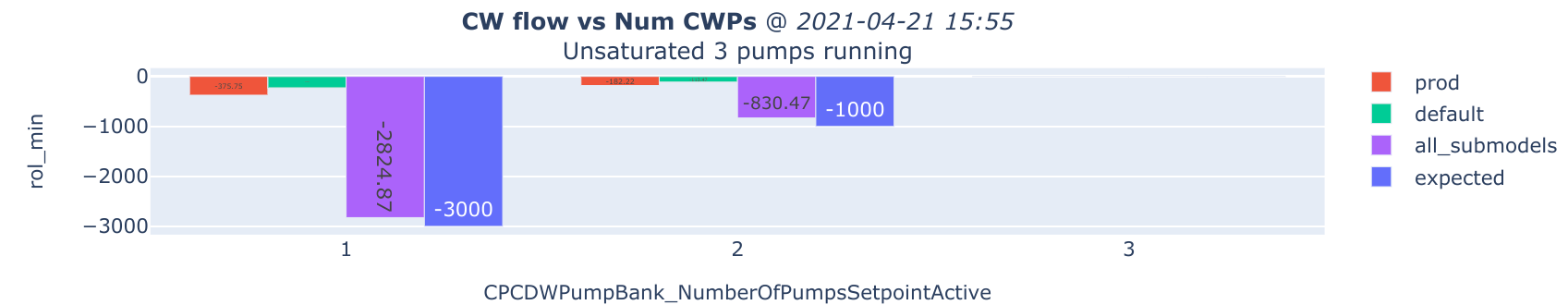}
\caption{A comparison of models on selected individual unit tests. We see that the model with the multi-tower architecture (labelled "all\_submodels") agrees much more with the expected results than the standard multi-headed model architecture (labelled "default").}
\label{fig:feature_selection}
\end{figure*}

The model's insensitivity to its action features strongly affected the agents performance (see Section \ref{section:challenge-limited-samples}). Therefore it was essential to detect this problem. Relying on simple metrics such as prediction MSE is insufficient for this task, so we built an evaluation framework based on sensitivity analysis of the model's predictions with respect to its features. Using this framework, we can compare the model's sensitivity to various actions with what we expect that sensitivity to be. 

The first step in this process is to deduce these expected sensitivities in various situations. This was done by a combination of deep domain understanding and analysis of the historical data to find the correct relationships. For example, we knew from domain experts that running too many chillers while the building load was low uses more energy than necessary. To verify the extent of this, we can isolate data points where the building load is low, among other confounders, and look at the general relationship between the number of chillers and energy consumption  (see Figure \ref{fig:power_vs_n_chillers}). This is then used as the expected value for this relationship. We eventually came up with a few dozen of these relationships, all supported by domain understanding and thorough analysis of the historical data.

Once the most important patterns were identified we constructed from them a set of individual unit tests for the model (see Figure \ref{fig:individual_unit_tests}). These involved selecting a particular data point and computing how the model predictions change as we change a single action feature with all other features fixed, and comparing it with how we expect the predictions should change. In the number of chillers example, we could start with a data point with two chillers, then change it to one chiller and see the model's response to this change. Then by comparing the observed behavior to the relationships extracted from data we can conclude whether the model has learned the relationship or not. During the earlier iterations of the agent the model did not react to these changes at all, and were completely ignoring the actions (as discussed in Section \ref{section:challenge-limited-samples}). Building this evaluation framework was crucial in identifying the issue, and allowed us to later resolve it (see Section \ref{section:submodels}).

\begin{figure*}[h]
    \includegraphics[width=\textwidth]{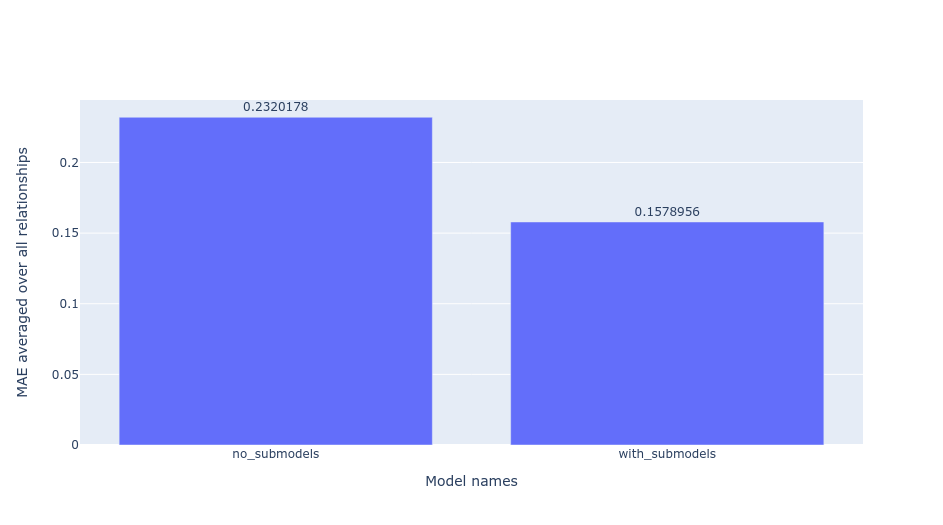}
    \caption{Example of an aggregated unit test comparing the performance of the model with and without the submodels architecture (See Appendix \ref{section:submodels}). These scores are computed as the averaged difference between the actual change in prediction vs the expected change in prediction across multiple individual unit tests. Thus, we have a single metric that we can use to compare different models that is more useful than the standard MSE loss, and also easier to use than the individual unit tests.}
    \label{fig:aggregated_unit_tests}
\end{figure*}

After several iterations we ended up with a large set of auto-generated individual model unit tests, so looking manually at each of them became intractable. To overcome this issue we decided to come up with one aggregated metric by computing the normalized MAE for each relationship, then take their average. $$metric = \mathbb{E}_{r \in Relationships} [\mathbb{E}_{d \in Data} \frac{|pred_{d,r} - actual_{d,r}|}{norm_{r}}]$$ For most relationships, we used the standard deviation of the ground truth trend as the normalization factor, but for the energy consumption objective, we used a manually chosen constant because its standard deviation was mostly attributed to the influence of weather. We called this the aggregated unit test, and it became the main deciding factor on whether to deploy a new model or not (see Figure \ref{fig:aggregated_unit_tests}).

\subsection{Feature engineering and submodels}
\label{section:submodels}

Once we identified that the model ignored the action features (see Section \ref{section:challenge-limited-samples}) and built the evaluation framework (see Section \ref{section:sensitivity-analysis}) to isolate the problem, we next had to force the model to learn from action features. This was achieved via feature engineering. Using domain understanding we started with a small set of actions that directly contributed to the prediction. These were supplemented with a handful of state features with high importance scores and that are known not to be significant confounders. To analyze feature importance we implemented the variance based approach for neural networks \citep{sa2019variance} \footnote{We also compared the obtained results with the standard analysis for random forest and observed a good agreement.}. Feature selection significantly improved model's performance on the unit tests (see Section \ref{section:sensitivity-analysis}). For example, we can see this in Figure \ref{fig:feature_selection}. As part of improving unit test performance through feature engineering, sometimes other metrics such as MSE increased, but we were willing to sacrifice this in exchange for the model's ability to understand the causal effect of actions. In total we selected 50 observations out of a total of 176 available to use.

The model had numerous different prediction heads that could all potentially depend on a different set of features. For example, the prediction for the flow rate may only need to depend on the flow rate setpoint and a few other state features that are adjacent to the water pump. However, the prediction for the energy consumption will need to consider almost every action. Because of this, we partitioned the prediction heads into sets that could share a set of input features, and did feature engineering for each set of prediction heads separately.

As a result of the feature selection, the critic architecture transformed from one multi-headed MLP into a multi-tower multi-headed MLP (see Figure \ref{fig:submodels_architecture}). This architecture is described in more detail in Section \ref{policy_evaluation_with_constraints}.

Further feature engineering, such as constructing more complicated non-linear features, did not show major improvements over a simple feature selection.

\subsection{Training on AI only data}
\label{section:training_ai_only}

Even with the multi-tower architecture (see Section \ref{section:submodels}), some important relationships that we expect  were still not learned properly. For example, the model was still not able to learn how the number of condenser water pumps affected total energy usage. Because these pumps are quite power-hungry it was important to be able to learn this relationship.

One solution to this was to only train on BCOOLER's exploration data, and ignore the data produced by the SOO policy (see Section \ref{sec:soo}). This is in spite of the fact that there was an order of magnitude more data from the SOO than BCOOLER. Our hypothesis is that since the SOO is a static policy, the strong correlations in the data didn't allow for the model to disentangle the effects of this action.

\subsection{Hierarchical heuristic policy}
\label{section:adding-heuristics}
As previously mentioned in Section \ref{section:challenge-multi-timescale}, some actions have longer lasting effects than others. In our case, one action with an exceptionally long lasting effect was changing the number of chillers that are running. In order to prevent the chillers from cycling on and off too frequently, one constraint was to make sure that the chillers don't turn on and off more than once every 2 hours. This means that before the agent can decide to turn off a chiller, it must be able to predict whether there is enough cooling provided by $N-1$ chillers for the next 2 hours. This is in contrast to the other actions that only need to plan anywhere from 5 minutes to 15 minutes ahead.

In order to make learning easier for the model, we ended up handling the number of chillers action separately using a heuristic, and delegating the rest of the decision-making process to the model. In this case, we can think of BCOOLER as a hierarchical agent, where there is a high level agent and a low level agent. The high level agent observes the state and uses it to make a decision about the number of chillers to run, and passes both its decision and the observation to the low level agent to make the decision about the rest of the actions. 

Using domain understanding of what should be safe number of chillers to run, and combining it with historical data analysis, we came up with a set of rules for the high level agent to control the number of chillers. The low level agent follows the core inference algorithm (see Section \ref{sec:policy_improvement}), and uses a model that only has a 15 minute prediction horizon.

\subsection{Tweaking the constraint specification}
\label{section:changing-constraints}
It is easy to determine if a particular situation violates some constraint, but difficult to enumerate all the possible constraints a priori (see Section \ref{section:challenges-constraints}). Before starting the live experiments, we worked with Trane and the facility managers to come up with a long list of action and observation constraints that the agent must satisfy. The facility managers have never thought about constraints in this way before, so this list was created specifically for BCOOLER to the best of everyone's knowledge. Because these constraints can have complex interactions with each other, the behavior of an agent trying to optimize its objective while satisfying these constraints didn't match our expectations. For example, there were numerous times when BCOOLER was overly constrained.

One way that this was discovered was that during some control iterations no actions were generated. When debugging the system we realized that after the rounds of filtering done during the control algorithm (see Section \ref{sec:policy_improvement}), no actions were left to select from. Upon deeper analysis, we narrowed down the set of constraints that were conflicting with each other, and worked together with Trane to fix the issue. 

Another way that this was discovered was by comparing BCOOLER's decisions with the SOO's decisions. This was something that was regularly done to make sure that BCOOLER was working properly, and also a way to analyze the strategies that BCOOLER used to optimize the system (see Appendix \ref{section:appendix_ai_learnings}). Several times during this analysis, we discovered that the SOO had taken actions that BCOOLER was not allowed to take due to constraint violations, and that BCOOLER would have taken these actions if it was allowed to. We worked together with Trane to relax the constraints.

Only through close collaboration with the HVAC domain experts were we able to fine tune the constraints such that the outcomes agreed with our expectations. This was a time consuming process, but was a necessary step to making sure BCOOLER is working well.

\subsection{Action pruning}
\label{section:action_pruning}

As mentioned in Section \ref{policy_evaluation_with_constraints}, due to the presence of constraints it was difficult to use a policy network to generate actions, and our algorithm performs a grid search instead. This search needs to be efficient, but at the same time cover enough of the search space to not miss important candidate actions. 

Because we have both continuous and discrete action dimensions, we first discretize the continuous dimensions in order to take their Cartesian product. In order to ensure good coverage of each continuous action dimension, we take a non-uniform random sampling of each continuous action dimension based on a log-uniform distribution with support between the smallest and largest allowed step size, both for increasing and decreasing the action's value from the previous timestep. This way more actions are generated that are close to the previous timestep's action to allow the agent to do fine tuning, but a few actions are generated further away from the previous action to allow the agent to make more coarse grained adjustments.

In order to handle the exponential complexity of the number of candidate actions to generate, we take a constraint satisfaction approach of pruning each dimension based on the relevant action constraints for that dimension before moving on to generate the next dimension. With our particular combination of action constraints, number of action dimensions, and values per dimension, we generate on the order of 100k actions. We cap the actual number of generated candidate actions to 100k by accepting each candidate action with a probability of $\min(1, 100000/N)$, where $N$ is the number of total candidate actions generated if there were no cap. This ensures that the downstream scoring and ranking algorithms only have a limited number of inputs to handle.

\subsection{Mode-specific action masking}
\label{section:action_masking}

As described in Section \ref{section:different_operation_modes}, the agent needs to be able to control the facility both while it is operating in mechanical cooling mode and while it is operating in free cooling mode, as well as be able to determine when to switch between the two modes. These modes have different action spaces and different constraints that are active when they are each enabled. To keep things simple, our agent always produced the superset of all actions available during both modes, and the actions that aren't relevant to a particular mode are ignored by the environment.

\subsection{Data examination and cleaning}
\label{section:data_cleaning}

During the beginning of the collaboration with Trane and the facility manager, we looked over both the historical data and current data from the facility to see whether they can be used for training. During this process we found several issues, including missing data, sensors reading physically implausible values, sensors having sudden jumps in their readings, sensors that are stuck reading the same values, sensors that have different units than the constants used in the constraint definitions, etc. Through careful analysis working together with Trane we were able to identify these issues. Some were able to be fixed via recalculating the value or inferring it from other variables. Others that couldn't be fixed were dropped from the training set. From discovering these issues in the first facility, we compiled a checklist to go through when looking over the data in the second facility, which made the process a lot quicker.

During inference, similar types of data issues also came up. Many of them are automatically detected by a set of checks in our system. For example, if a relevant observation isn't found in the current 5 minute window, then a configurable look-back period is used to fill in the missing measurement in the cases where there is a small gap in the measurements for a particular sensor. In other cases, the issues were only identified after closely monitoring BCOOLER's performance, or observing that BCOOLER was kicked out (see Appendix \ref{sec:appendix_ai_safety}). The team built tools to debug these data issues, and we worked closely together with Trane and the facility manager to resolve them.

\section{Related work}

\paragraph{Reinforcement learning for HVAC control:} There has been significant interest in using RL approaches to HVAC control - see \citet{yu2021review} for a recent comprehensive overview. However, most of these works are only evaluated in simulation. Notable exceptions include \citet{chen2019gnu}, who control a single room in an office building, \citet{zhang2019whole}, who control one floor of an office building, and \citet{kazmi2018gigawatt}, who control the hot water tanks and heat pumps for 20 houses. The data center optimization work \citep{google-ai-data-center-2} that BCOOLER was based on is a very similar controls problem, though BCOOLER adds additional actions such as controlling the number of chillers and number of pumps. The consequences of adding these additional actions is that BCOOLER also handles significantly more constraints. \citet{nweye2022offline} also describe how the challenges in \citet{challenges} relate to RL for HVAC control, however this is not matched to a case study of deploying on a real system.

\paragraph{Offline reinforcement learning:}
\label{sec:offline_rl}
In our case, as well as for many other applied RL problems, it is important for the agent to be able to train using offline data logs (see Appendix \ref{section:challenge-limited-samples}). Research in offline RL has been progressing a lot in recent years - see \citet{levine2020offline} for an overview. A natural approach to offline RL is to downgrade the value functions for state-action pairs that have not been seen very often in the offline dataset \citep{kumar2020conservative,gulcehre2021regularized}. Our approach of penalizing by the standard deviation of the Q value ensemble can be seen as an example of this type of approach, since state state-action pairs that are frequently seen in the dataset should have lower ensemble uncertainty than rarely encountered ones.

\paragraph{Exploration and regularization via ensemble uncertainty:} BCOOLER trades off between exploration and behavior regularization\footnote{By behavior regularization we mean encouraging the learned policy to stay close to the behavior policy that generated the offline dataset.} based on the standard deviation of the value function ensemble. While we did not find an instance of the exact same algorithm in the literature, there are examples of similar approaches for either only exploration or only behavior regularization. The exact approach of using a value function bonus based on the ensemble standard deviation is taken by \citet{chen2017ucb} (see Algorithm 2). Similar approaches can be found in several other works \citep{osband2016uncertainty, buckman2018sample}. Examples of behavior regularization by using an ensemble standard deviation penalty for the value function can be found in \citet{an2021uncertainty} and \citet{ghasemipour2022so}. Other approaches that penalize the value function in order to encourage the agent to stay close to the behavior data distribution include \citet{kumar2020conservative} and \citet{gulcehre2021regularized}.

\section{Conclusion}

In this paper we outline the work that was done applying RL control to multiple commercial HVAC facilities. Through this process we uncovered several challenges associated with a real world industrial controls problem that aren't present in most simulated and virtual environments. We found that no off-the-shelf RL algorithm can work in a plug-and-play fashion, and made significant modifications to a simple RL algorithm in order to solve the challenges faced. Significant domain understanding was required for many of the modifications made. As a result of all this, our control algorithm was able to demonstrate around 9\% and 13\% savings during live testing in two real world facilities.

\paragraph{Future work:}
There are several directions that we can focus on to improve BCOOLER. Data efficiency could be improved by adding additional domain specific inductive biases to the model, such as the physical topology of the equipment in the facility, or the constraints and invariances of the sensor measurements given by first principle physics.  Another idea would be to build simulations of different facilities with a variety of attributes, such as equipment type, equipment topology, environmental conditions, etc. and use this to transfer knowledge to agents that will be deployed to real facilities. This can range from simpler ideas like sanity checking model changes in simulation, to more advanced techniques such as meta and transfer learning or even direct sim-to-real transfer. This line of simulation work has started to be explored in \citet{industrial_task_suite}. Another direction is to focus on the generalizability of the algorithm, because large scale impact requires deployment to new facilities without significant engineering, modeling, and problem definition work per facility. For instance we could construct better representations of the facility ontology to make it easier to translate the tasks that customers have in mind to a language that the agent can understand. Another idea would be to more effectively leverage direct human feedback to streamline the user experience of teaching the agent to behave correctly for facilities of various types.

\vspace*{\fill}

\pagebreak

\bibliography{main}

\clearpage
\appendix

\section{Algorithm details}

\subsection{Policy improvement with constraints details}
\label{sec:appendix_policy_improvement_details}

The policy improvement phase, also referred to as control or inference, is when the learned action value function from the policy evaluation phase is used inside a policy.  The policy here isn't defined by a policy network. Instead the learned action value function is used to search the space of possible actions. The high level goal of the policy is to utilize both the facility configuration and the learned action value function in order to map the current sensor observations into an action that will minimize expected energy usage over time, while satisfying both action and observation constraints.

The first step in the policy is candidate action generation, where we take the Cartesian product of all action dimensions by discretizing or sampling the continuous dimensions of the action space. The combination of candidate actions is exponential with respect to the action dimensions and is highly sensitive to the number of values per dimension. Without action constraints, there are on the order of 10M actions using a coarse discretization of 4 values per continuous dimension. With action constraints, the enumeration of candidate actions becomes much faster in our case, typically generating on the order of 100k actions. We use intelligent action pruning to ensure that the most relevant actions are generated, as detailed in Section \ref{section:action_pruning}.

Next we attach the current observation to each candidate action and feed the result to the action value function, which will give the objective and observation constraint predictions for each candidate action. Using the observation constraint predictions, together with the observation constraint inequalities defined in the facility configuration, we can filter out the candidate actions that are predicted to violate the observation constraints. In order to reduce the likelihood of violating observation constraints, the agent uses the standard deviation of the ensemble predictions to be pessimistic with respect to the observation constraint. In other words, the higher the standard deviation of the observation constraint prediction, the further away the predictions need to be from the constraint boundary in order to not get filtered out.

Finally we need to choose an action from the remaining candidate actions. The agent uses an optimistic modification of the $\epsilon$-greedy algorithm for exploration, where most of the time the agent exploits by picking the best action according to its internal models (i.e. exploitation), and 5\% of the time the agent will explore an action that looks promising (i.e. exploration). 

During exploitation, the standard deviation in the ensemble prediction will be used to penalize the action's predicted return. Then the best action with respect to the adjusted predicted return is selected. This will promote the agent picking good actions that it's confident about. 

During exploration, the standard deviation in the ensemble prediction will be used to reward the action's predicted return \citep{osband2016uncertainty}. Then the adjusted predicted return will be used to randomly choose from the remaining candidate actions, where the probability for choosing action $a_i$ is

\begin{equation}
    P(a_i) = \frac{e^{\beta \cdot \textrm{score}_i}}{\sum_j e^{\beta \cdot \textrm{score}_j}}
\end{equation}

where $\textrm{score}_i$ is the adjusted predicted return for action $a_i$, and $\beta$ is a parameter that can be adjusted to control the exploration entropy. This will promote the agent to pick actions that it is uncertain but also optimistic about.

If for some reason the combination of constraints results in an empty set of candidate actions, then the agent will relinquish control back to the SOO, as described in Appendix \ref{sec:appendix_ai_safety}.

\begin{center}
\begin{table*}
\centering
\begin{tabular}{||c c c||} 
 \hline
 Action name & Range & Units \\ [0.5ex] 
 \hline\hline
 Chiller 1 temperature & [40, 52] & $^\circ$F \\ 
 \hline
 Chiller 2 temperature & [40, 52] & $^\circ$F \\
 \hline
 Chiller 3 temperature & [40, 52] & $^\circ$F \\
 \hline
 Number of mechanical cooling chillers & \{0, 1, 2, 3\} & None \\
 \hline
 Number of cooling towers & \{1, 2, 3\} & None \\
 \hline
 Cooling tower temperature & [40, 100] & $^\circ$F \\
 \hline
 Condenser water pump flow & [700, 5400] & gpm \\
 \hline
 Number of condenser water pumps & \{1, 2, 3\} & None \\
 \hline
 Chilled water differential pressure & [12.5, 20] & psid \\
 \hline
 Number of chilled water pumps & \{1, 2, 3\} & None \\
 \hline
 Number of free cooling chillers & \{0, 1, 2, 3\} & None \\
 \hline
 Free cooling chilled water temperature & [42, 52] & $^\circ$F \\
 \hline
\end{tabular}
\vspace{.2cm}
\caption{The action space of BCOOLER.}
\label{tab:action_space}
\end{table*}
\end{center}

\subsection{Action space}
\label{section:appendix_action_space}

As described in Section \ref{section:rl_problem_formulation}, there is a mixture of continuous and discrete dimensions in the action space. Table \ref{tab:action_space} shows the dimensions in more detail for one of the facilities. In reality, the continuous dimensions can't actually take any arbitrary real number in the range, since the underlying equipment's precision doesn't allow for that granular of control. In practice, the facility configuration defines a step size $s$ for each dimension and the agent must make changes of size $ns$, where $n$ is an integer.

\subsection{AI safety}
\label{sec:appendix_ai_safety}

By default the SOO fully controls the facility. Once BCOOLER is ready it starts sending recommendations to the facility, but the facility will only start accepting those recommendations when 
the BMS operator flips a switch on their end to enable AI control. AI control can be disabled at any time to revert to SOO control for a few reasons. One automated reason is that BCOOLER has violated some constraint by more than the tolerable amount. Another automated reason is that BCOOLER failed to generate any actions due to a bug in the system or conflicts in the constraints. 
We (as well as the BMS provider) also had the option to manually disable BCOOLER if either party was uncomfortable with its recommendations at any point. In this case, Trane was once again required to manually flip the switch in order to resume AI control.

\subsection{Facility configuration}
\label{section:appendix_facility_configuration}

This RL controls system is meant to be a easily adaptable to many environments, and in order to configure it we define a facility configuration that is tailored to each facility. This configuration defines the set of sensors that the agent can observe, the actions that the agent will have access to, the objective that the agent is optimizing for, and the constraints that the agent must satisfy.

The constraints are defined in the facility configuration as a set of inequalities containing actions and/or observations. For example, a simple action constraint can be $a_1 \leq 42$, where $a_1$ corresponds to the temperature setpoint of one of the chillers. The action constraints can be met with certainty since we can guarantee that the agent doesn't recommend an action that violates the set of action constraints. A simple observation constraint can be $\max (\theta_1, \dots \theta_{T_c}) \leq 45$, where $\theta_t$ corresponds to the value of a specific temperature sensor at timestep $t$. Even though this observation constraint is simple to state, it can be difficult to achieve because actions can have long lasting effects, which means an agent must be able to plan ahead in order to avoid violating it. Combined with the fact that there are stochastic elements to the environments such as weather, it may not be possible to guarantee that observation constraints are always satisfied.

To simplify the work of the agent, we added the restriction on the facility configuration that for any observation constraint the constraint function $c_s$ in Equation \ref{eq:main_objective} corresponds to a specific sensor (also referred to as the "constrained observation") that can be affected by the agent's actions and that will have constraints on it, whereas the bounds correspond to a function of constants and observations that define the boundary of the constrained observation.

\subsection{Hyperparameters}
\label{section:appendix_hparams}

Our action value function Q is represented by a neural network with 1 shared hidden layer with 2 additional hidden layers per head, and has 128 units for each hidden layer (See Figure \ref{fig:submodels_architecture}). We use 10 models with identical architecture but different initialized weights in our ensemble. The prediction horizon for the energy objective and observation constraints is 15 minutes. During training we use an Adam optimizer \citep{adam_optimizer} with a learning rate of 0.001. During control, we set an exploration rate $\epsilon$ of 0.05, an ensemble standard deviation penalty $\alpha$ of 1, and an exploration temperature $\beta$ of 0.01 (See Appendix \ref{sec:appendix_policy_improvement_details}).

\section{Examples of improvements discovered by BCOOLER}
\label{section:appendix_ai_learnings}

After analyzing the decisions that BCOOLER has made and comparing it to the baseline, we can see some interesting behaviors that it has found to improve the efficiency of the control system.

\subsection{Balancing power usage between chillers and cooling towers}
\begin{figure*}[h]
    \begin{center}
    \includegraphics[width=0.9\textwidth]{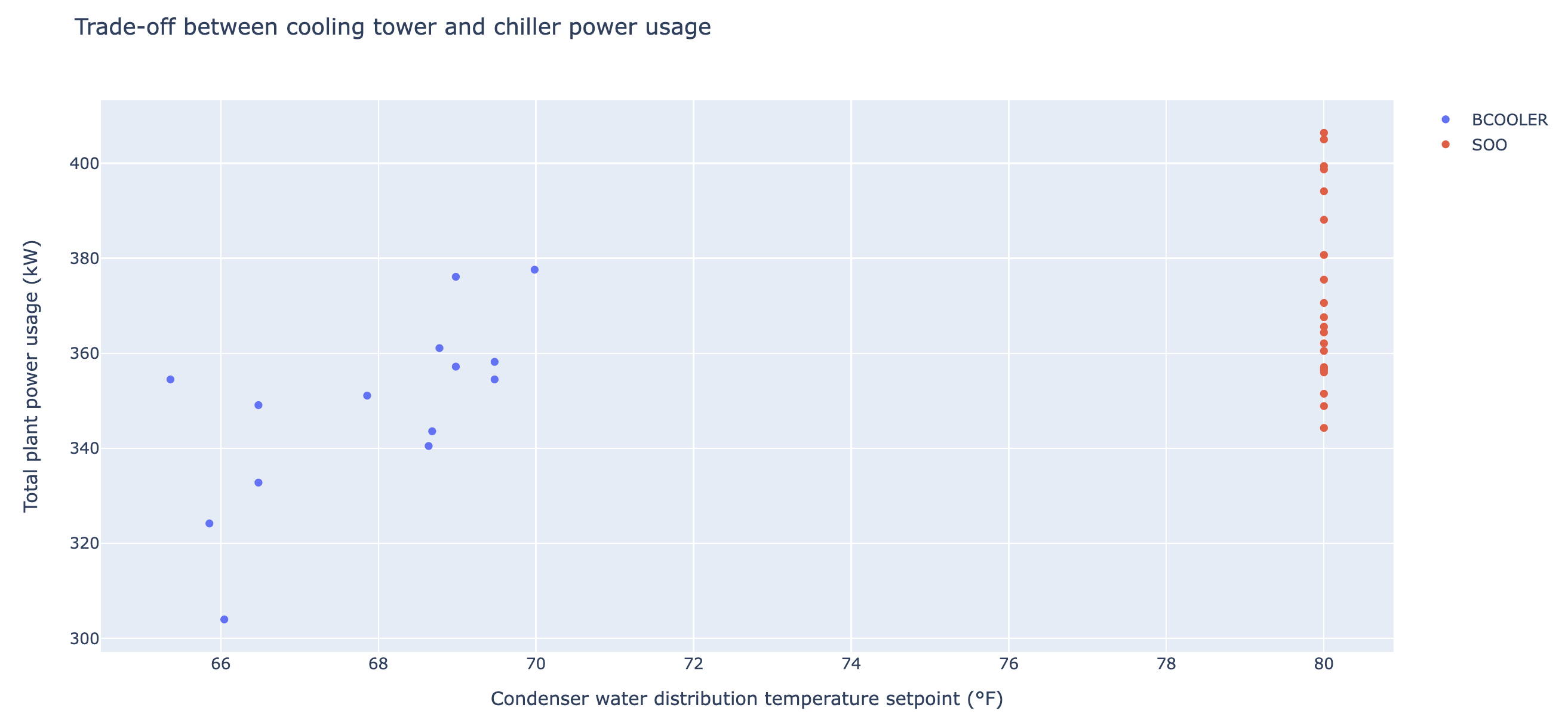}
    \end{center}
    \caption{Setting the condenser water temperature dictates how much power is used between the cooling tower and the chiller. Setting a lower condenser water temperature means that the cooling towers work harder to produce colder condenser water, but this makes the chiller's job easier since it will then need to transfer energy to colder water. We see that the SOO's choice for the condenser water temperature setpoint is static, and that BCOOLER chooses on average lower setpoints than the SOO, which correlates with lower total plant power usage. We try to account for confounding factors by only selecting a few points that have similar values for the outside wet bulb temperature, building load, the number of chillers running, and the temperature of the distribution supply temperature.}
    \label{fig:condenser_chiller_balance}
\end{figure*}

One optimization that BCOOLER found was to balance the power usage between the chiller and the cooling tower. These are two pieces of equipment that compete for power usage. If more power is put into the cooling tower, then less power is required for the chiller, and vice versa. The optimal trade-off depends on a variety of factors including the current load, outside air temperatures, humidity, and the number of chillers and cooling towers running. Due to the complex interaction between all of these variables, the SOO is programmed with simple coarse rules on how to make this trade-off. However, BCOOLER is able to learn from historical data in order to make more nuanced decisions, and we observed that its recommendations were oftentimes significantly different from the SOO (See Figure \ref{fig:condenser_chiller_balance}).

\subsection{Accounting for sensor miscalibration}
Another optimization that BCOOLER found was to correct for the fact that certain sensors fell out of calibration. For example, there is a master supply water temperature sensor used to measure whether the temperature constraints have been violated, but a separate evaporator leaving water temperature sensor inside each chiller used to control to its water temperature setpoint. These sensors should read about the same in many circumstances since they are measuring the water temperature at nearby points, but due to sensor miscalibration they could read very differently (see Figure \ref{fig:sensor_drift} for an example). There were instances where the agent was trying to maximize the master supply water temperature, and the upper limit for this temperature's constraint was  45$^\circ$F. Also, the chiller evaporator leaving temperature read 2$^\circ$F higher than the master supply water temperature, which means a chiller with a 45$^\circ$F setpoint will actually achieve 43$^\circ$F water as measured by the master supply temperature sensor. BCOOLER was be able to understand that it is okay to set the chiller setpoint to 47$^\circ$F, because due to the miscalibration the master supply temperature will end up being 45$^\circ$F, which is within the limits of the constraint.

\end{document}